\documentclass[11pt,letterpaper]{article}
\usepackage[T1]{fontenc}
\usepackage[utf8]{inputenc}
\usepackage{times}
\usepackage[letterpaper,margin=1in,headheight=14pt]{geometry}
\usepackage[table]{xcolor}
\usepackage{amsmath,amssymb,bm}
\usepackage{graphicx,booktabs,tabularx,multirow,array}
\usepackage{microtype}
\usepackage{enumitem}
\usepackage[section]{placeins}
\usepackage{caption}
\usepackage[numbers,sort&compress]{natbib}
\usepackage{xspace}
\usepackage{titlesec}
\usepackage{fancyhdr}

\definecolor{linkblue}{HTML}{2D5878}
\usepackage[breaklinks,colorlinks,allcolors=linkblue,bookmarksopen=false]{hyperref}
\usepackage[capitalize,nameinlink,noabbrev]{cleveref}

\newcolumntype{L}{>{\raggedright\arraybackslash}X}
\newcolumntype{C}{>{\centering\arraybackslash}X}
\newcommand{\method}{\texorpdfstring{A\textsuperscript{2}-VLA}{A2-VLA}\xspace}

\setlist[itemize]{leftmargin=1.4em,topsep=4pt,itemsep=3pt,parsep=0pt}
\setlist[enumerate]{leftmargin=1.6em,topsep=4pt,itemsep=3pt,parsep=0pt}
\titleformat{\section}{\large\bfseries}{\thesection}{0.7em}{}
\titleformat{\subsection}{\normalsize\bfseries}{\thesubsection}{0.7em}{}
\titleformat{\paragraph}[runin]{\normalsize\bfseries}{}{0pt}{}
\titlespacing*{\section}{0pt}{17pt plus 3pt minus 2pt}{7pt}
\titlespacing*{\subsection}{0pt}{12pt plus 2pt minus 2pt}{5pt}
\titlespacing*{\paragraph}{0pt}{8pt plus 2pt minus 1pt}{0.7em}

\fancypagestyle{plain}{%
 \fancyhf{}\fancyfoot[C]{\small\thepage}%
 }

\makeatletter
\renewcommand{\maketitle}{%
 \begingroup\centering
 {\Large\bfseries\@title\par}
 \vspace{8pt}
 {\normalsize\@author\par}
 \vspace{3pt}
 {\normalsize\authoraffiliation\par}
 \vspace{3pt}
 {\footnotesize\authornote\par}
 \vspace{5pt}
 {\small\textsc{Preprint}\par}
 \vspace{8pt}
 \endgroup\thispagestyle{plain}}
\makeatother
\renewenvironment{abstract}{%
 \par\vspace{7pt}\noindent
 \begin{minipage}{\linewidth}
 \begin{center}\bfseries Abstract\end{center}
 \begin{list}{}{\setlength{\leftmargin}{18pt}\setlength{\rightmargin}{18pt}}
 \item[]\small
}{\end{list}\end{minipage}\par\vspace{8pt}}

\title{A\textsuperscript{2}-VLA: Adapting Vision-Language-Action Policies\\
to New Tools through Agentic Affordance Discovery}
\author{Chenyuan Zhou\textsuperscript{*}\quad Zibo Wang\textsuperscript{*}\quad Chaoyi Huang\textsuperscript{*}\quad Zhiyuan Li\quad Yong-Lu Li\textsuperscript{\dag}}
\newcommand{\authoraffiliation}{Shanghai Jiao Tong University}
\newcommand{\authornote}{\textsuperscript{*}Co-first authors.\quad\textsuperscript{\dag}Corresponding author.}
\date{}
\hypersetup{
 pdftitle={A2-VLA: Adapting Vision-Language-Action Policies to New Tools through Agentic Affordance Discovery},
 pdfauthor={Chenyuan Zhou; Zibo Wang; Chaoyi Huang; Zhiyuan Li; Yong-Lu Li},
 pdfsubject={Preprint},
 pdfkeywords={Tool Adaptation, Affordance Discovery, Vision-Language-Action Models, Embodied Agents, Simulation-Guided Refinement},
 pdfcreator={LaTeX}
}

\begin{document}
\maketitle
\begin{minipage}{\textwidth}
\centering
\captionsetup{hypcap=false}
\includegraphics[width=0.97\linewidth]{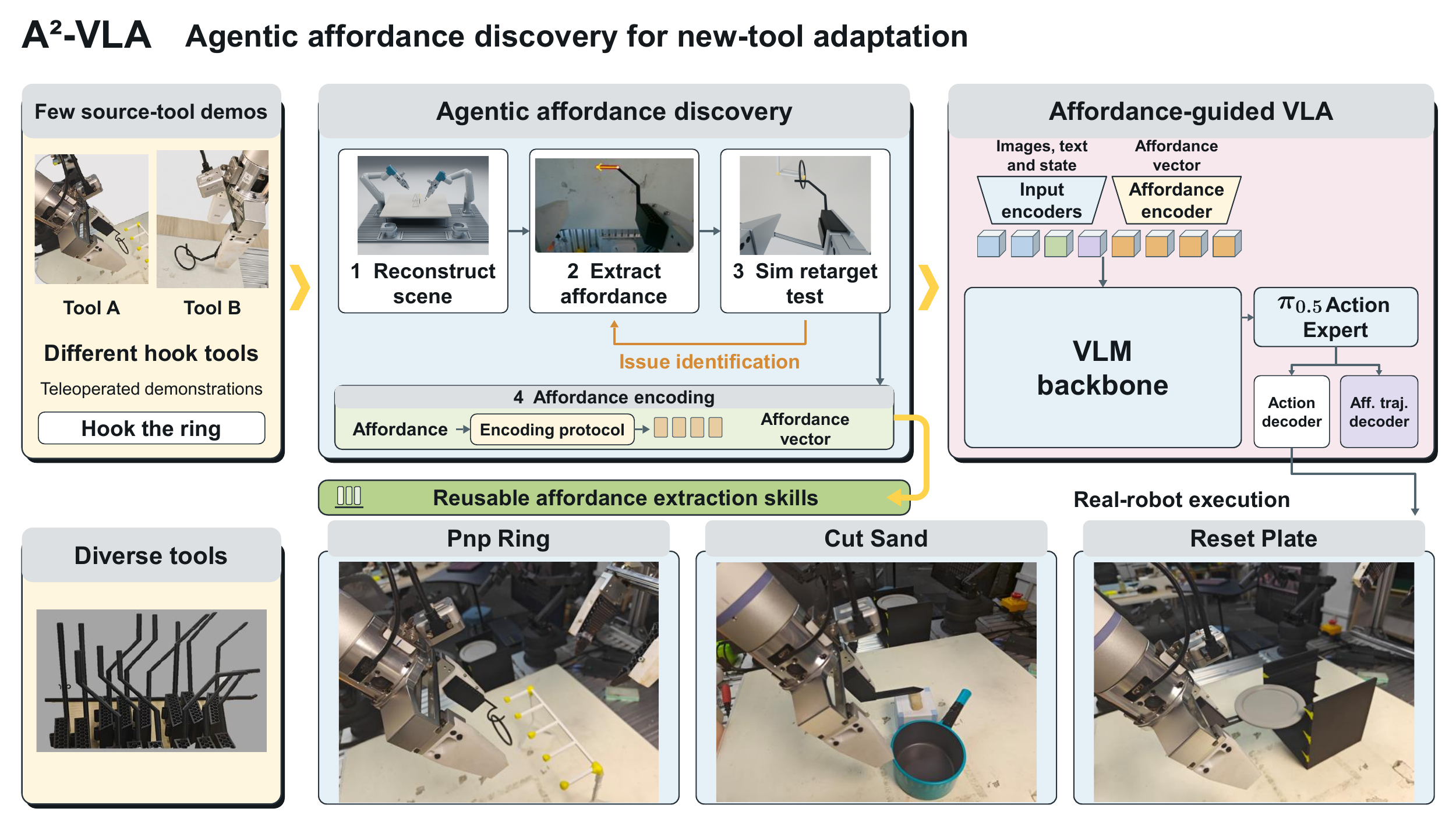}
\captionof{figure}{Given demonstrations of a task performed with a few source tools, \method analyzes
functional-region features, reconstructs scenes in simulation, and validates the inferred affordances.
Refined extraction skills annotate the demonstrations with affordance features for VLA training.
For an unseen tool, \method uses validated skills to extract its affordance, verifies it in simulation,
and supplies the encoded functional-region features to guide VLA execution.}
\label{fig:overview}
\end{minipage}
\vspace{5pt}

\begin{abstract}
Tools with different shapes can have similar functions, but transferring manipulation skills
across tools requires understanding how tool geometry relates to the specific task and adjusting
the manipulation strategy accordingly. The key challenge is to discover the common affordances
that contribute to task completion across different tools and verify whether they support new
tools in performing similar work. We present \method, a tool-adaptation framework that combines
an embodied agent with a VLA policy. It treats tool affordances as hypotheses the agent can construct,
test, and revise. Given teleoperated demonstrations with a few source tools, the agent
reconstructs the scenes in simulation and autonomously determines task-relevant affordance
features: functional regions and their key geometric properties. It develops
affordance-extraction skills and refines both the features and skills through autonomous
experiments in the reconstructed scenes. Validated extraction skills accumulate in a reusable
skill library for unseen tools. Finally,
\method uses encoded functional-region features as conditioning inputs to a $\pi_{0.5}$ backbone during
training and inference. Evaluation examines affordance reliability in reconstructed scenes and
real-robot task success on unseen tools. Our method achieves 86.1\% success
in simulated cross-tool trajectory retargeting based on affordances.
On three real-robot tool-use tasks, \method improves task success by 30.7 percentage
points over $\pi_{0.5}$.

\end{abstract}

\section{Introduction}
\label{sec:introduction}

People can often reuse a familiar manipulation skill with an unfamiliar tool after inspecting it
and attempting to interact with the environment using it. A differently shaped implement may still provide
functional regions similar to those of a familiar tool, such as an edge for cutting or a tip for
hooking, even though its appearance and required wrist motion differ substantially from those of
the original tool. What transfers is therefore neither the tool's identity nor a fixed
end-effector trajectory, but an understanding of the properties that make the intended interaction
possible. Enabling robots to acquire and reuse this understanding from limited tool experience
would reduce the need to collect demonstrations for every new tool--grasp combination.

Toward this goal, we study how to enable a robot system to adapt to new tools that can perform
the same task, starting from teleoperated demonstrations using a small set of source tools.
Vision-language-action (VLA) policies already possess some tool-manipulation
capabilities~\cite{openvla,pi0,pi05}, but adapting them to entirely new tools still requires
repeated rollout collection and fine-tuning to maintain performance. Predefined functional
representations of tools have shown useful structure for transfer: MimicFunc establishes
functional correspondences through keypoints and
function frames~\cite{mimicfunc}, and FuncBridge learns functional trajectories separately from
their grounding into actions~\cite{funcbridge}. Meanwhile, improvements in the embodied
manipulation capabilities of general-purpose large vision-language models (VLMs) make autonomous
analysis of tool function possible. Our focus is how an agent system can autonomously identify
task-relevant functional regions and determine their key geometric features, capturing functional
commonality across tools to guide a VLA policy in adapting to new tools.

We treat task-relevant affordance features as experimentally testable hypotheses. Although
hierarchical robot systems already connect reasoning by general-purpose VLMs to low-level policy
execution through language or spatial guidance~\cite{hirobot,hamster,steerable}, language or
keypoints alone cannot fully represent the geometric details of tools. Cross-tool adaptation
additionally requires deciding which tool properties should be represented for a specific task
and testing whether these properties can guide the policy to complete it correctly. We introduce
\method, in which the agent starts from a few real-robot teleoperated demonstrations, reconstructs
the demonstrated scenes and tools in simulation, and conducts experiments within them.
The agent analyzes how different source tools accomplish the task, determines task-relevant
affordance features, develops affordance-extraction skills, and applies them to other tools to
attempt task execution. Based on experimental feedback, it analyzes unsuccessful interactions
and refines the selected functional regions, geometric features, or extraction skills.
Successful skills and their applicability conditions accumulate during this process in a
reusable skill library. When faced with an unseen tool, the agent can therefore reuse previously
validated skills to identify corresponding functional regions and extract the same features.

We condition the low-level VLA execution policy on agent-encoded functional-region
features through a low-dimensional affordance interface. While analyzing the source
tools, the agent determines task-relevant affordance features, which are encoded under a
unified protocol as conditioning inputs for VLA training. Unseen tools'
affordances are described through the same protocol: their specific affordance parameters are
extracted while the coordinate representation and feature semantics remain unchanged. This gives
the agent freedom to select functional regions and geometric features while maintaining
a consistent policy input format. Encoded features serve as conditioning inputs
when training a $\pi_{0.5}$ policy with an auxiliary affordance prediction head. During
deployment, the agent reconstructs the unseen tool in simulation, extracts and tests candidate
affordances, and refines their parameters from simulation feedback before encoding them
under the same protocol for the VLA policy.
At execution time, the VLA conditions each action chunk on the updated
affordance features and current observations.

Our evaluation examines the reliability and reusability of autonomously discovered
affordances and their effectiveness in cross-tool transfer.
Simulation tests compare our affordance discovery and validation process with
other tool-adaptation methods, including task success when using affordances to retarget
trajectories in reconstructed scenes. Three
real-robot tool-use tasks test the robustness of cross-tool transfer. Human Oracle uses human-provided
affordances for both VLA policy training and inference to distinguish failures caused by poor
affordance features from those caused by the VLA's execution capability.

Our main contributions are:
\begin{itemize}
  \item An agentic tool-adaptation framework that starts from teleoperated demonstrations using
  a small set of source tools and autonomously determines task-relevant affordance features.
  The agent reconstructs the demonstrated scenes in simulation, tests and refines these features
  through autonomous experiments, and retains refined affordance-extraction
  procedures as reusable skills across tools.
  \item An affordance-guided VLA model that integrates the resulting low-dimensional
  feature vectors into action generation. Affordance conditioning and auxiliary
  affordance-trajectory supervision train the policy to faithfully follow the supplied affordance
  information while using current observations
  and robot state to complete tool-use tasks.
\end{itemize}

\section{Related Work}
\label{sec:related_work}

\paragraph{Tool adaptation.}
Tool adaptation requires understanding how a tool itself functions in a specific task.
KETO~\cite{keto} and GIFT~\cite{gift} learn sparse grasp and functional keypoints, while
FuncBridge~\cite{funcbridge} predicts 2D functional keypoint trajectories to guide action
generation. However, functional points alone may not capture more complex manipulation
characteristics. Other work, such as MimicFunc~\cite{mimicfunc}, automatically extracts keypoints and
constructs function frames for cross-tool transfer. Its affordance schema and frame-construction
rules are predefined, however, so adaptation remains organized around interactions expressible
within that abstraction rather than autonomously determining which functional features a task requires.
MAGIC~\cite{magic} matches DINOv2 visual features and local curvature to transfer a demonstrated
contact strategy, with physical simulation checking candidate motions. Its candidate generation
relies on visual--geometric analogies and struggles with substantially different tools.
Other work uses simulation environments to discover and validate reliable tool geometries,
including GeT-USE~\cite{getuse} and ToolAnalogy~\cite{toolanalogy}, thereby improving the ability
to select suitable tools.

\paragraph{Vision-language-action models.}
VLAs learn observation-conditioned manipulation from multimodal pretraining and robot
demonstrations. Representative models include OpenVLA~\cite{openvla}, the flow-matching policy
$\pi_0$~\cite{pi0}, and $\pi_{0.5}$~\cite{pi05}. However, their manipulation knowledge remains
dependent on the behaviors and interactions covered by training data. Unfamiliar tool geometries
and contact patterns can expose gaps in this coverage, limiting cross-tool generalization.
Hierarchical architectures decouple high-level task-related information from low-level
policies, allowing task reasoning and action control to be modeled separately.
Hi Robot uses language subgoals and HAMSTER uses 2D paths; these intermediate interfaces
guide low-level execution with task goals or spatial plans~\cite{hirobot,hamster,steerable}.
Existing work also adds functional guidance to VLAs and related visuomotor policies,
grounding task intent in spatial interaction structure. RT-Affordance~\cite{rtaffordance}
uses key-stage end-effector poses, FuncBridge~\cite{funcbridge} predicts functional keypoint
trajectories, and AFI~\cite{afi} uses 3D affordance fields to guide waypoints and select
trajectories. These representations provide explicit cues about where and how to interact,
connecting semantic goals to geometrically grounded action generation.

\paragraph{Embodied agents.}
Embodied agents use general knowledge to analyze tool functions and formulate execution
strategies. ReKep~\cite{rekep} generates relational keypoint constraints, and
ToolAnalogy~\cite{toolanalogy} proposes functional features. Meanwhile, agentic real-to-sim
systems, such as WorldMirror and SceneRig, reconstruct interactive environments for trajectory
replay and policy testing~\cite{worldmirror,scenerig,agenticrsr}, allowing agents to test their
understanding of manipulation through interaction. For reusable skill development, ASPIRE~\cite{aspire} uses
execution feedback to repair control programs and distill reusable skills. \method applies
real-to-sim interaction testing and reusable skill development to tool adaptation. The agent
autonomously identifies task-relevant functional regions and determines their key geometric
features, then refines these features and extraction skills through simulation
feedback to guide VLA execution with unseen tools.

\section{Problem Formulation}
\label{sec:problem}

Given a task instruction $\ell$, we are provided a small set of source tools,
each with real-robot teleoperated demonstrations. At time $t$, the observation
$\mathbf o_t=(I_t,\mathbf q_t)$ contains camera images $I_t$ and the robot
configuration $\mathbf q_t$. The first command predicted from $\mathbf o_t$
is $\mathbf a_t$.
An action chunk $\mathbf A_t=(\mathbf a_t,\ldots,\mathbf a_{t+H-1})$
contains $H$ consecutive robot commands. The demonstrations provide the observations and trajectories
needed to reconstruct tool interactions in simulation.

We seek transfer to unseen tools, disjoint from source tools, without additional
task demonstrations. The agent autonomously determines task-relevant affordance features
by analyzing source-tool demonstrations, identifying functional regions and key geometric
properties shared across tools. It develops skills to extract corresponding features
from unseen tools. We denote the encoded functional-region feature vector at time $t$
by $\mathbf f_t$. Feature values follow a unified task-level encoding protocol.
For an unseen tool, the skills use inspection observations and its attachment to the robot.
Different tools supply different values under the same feature definitions and encoding.
The policy $\pi_\theta$ uses $\mathbf f_t$ together with
$\mathbf o_t$ and $\ell$ to complete the task. We evaluate affordance reliability
through simulated cross-tool retargeting and transfer performance through
real-robot task success.

\section{\method}
\label{sec:method}

\begin{figure}[tbp]
\centering
\includegraphics[width=\linewidth]{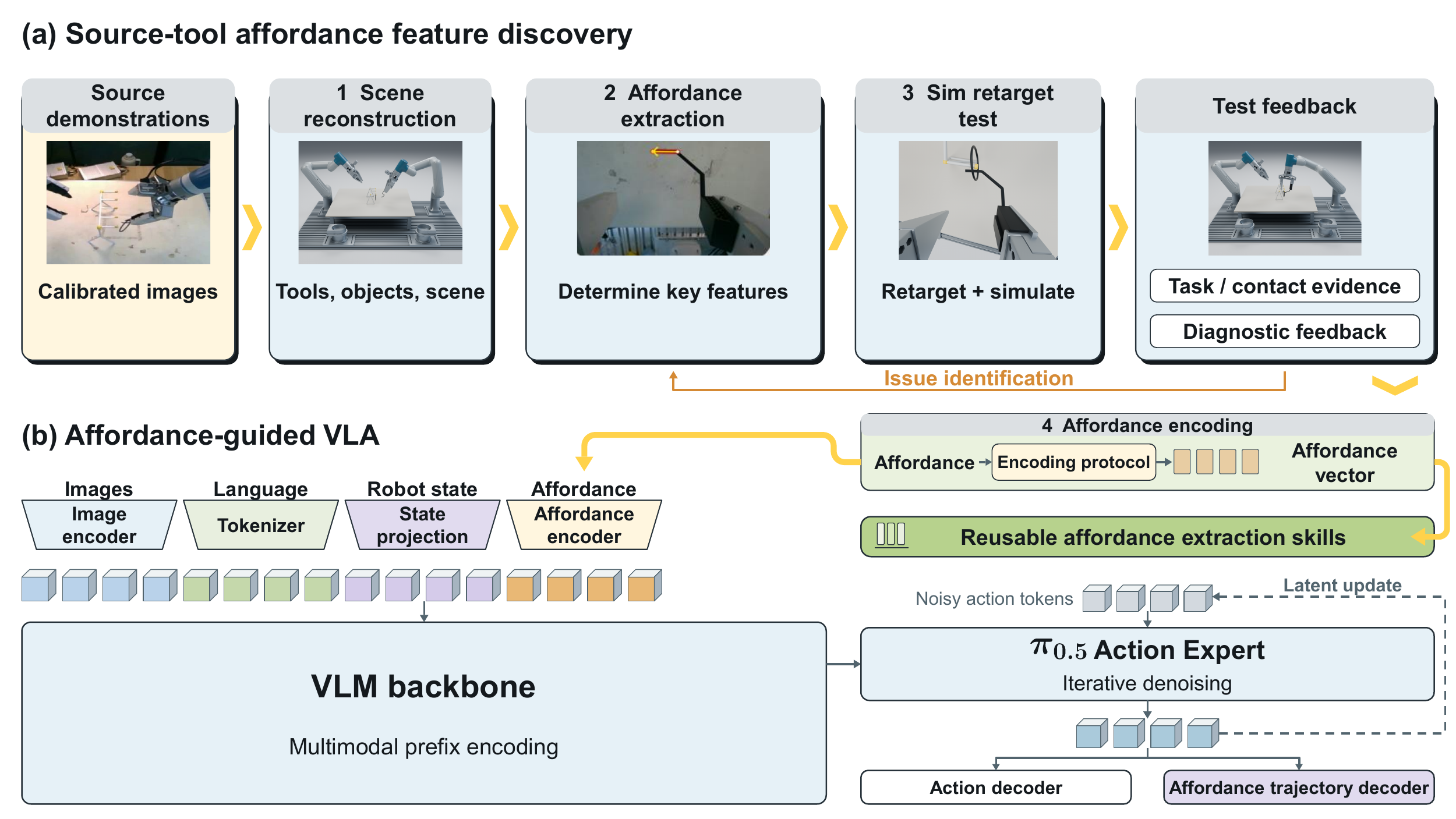}
\caption{\method overview. (a) The agent reconstructs demonstrated scenes, autonomously determines
task-relevant affordance features, develops affordance-extraction skills, and tests cross-tool retargeting
in simulation. Diagnostic feedback guides revisions; refined extraction skills enter the reusable library.
(b) Encoded functional-region features join image, language, and robot-state tokens to condition the
$\pi_{0.5}$ action expert. It iteratively denoises action tokens, decoded into robot
actions and an auxiliary affordance trajectory.}
\label{fig:method}
\label{fig:agentloop}
\label{fig:policy}
\end{figure}

\method connects an agent that analyzes tool function with a VLA that executes the
task (Fig.~\ref{fig:method}). The agent reconstructs demonstrated interactions,
develops affordance-extraction skills, and tests them through simulation.
Validated affordances condition VLA training and deployment.

\subsection{Reconstructing Demonstrated Interactions}
\label{sec:reconstruction}

The agent performs real-to-sim reconstruction using observations and recorded robot trajectories from source-tool
demonstrations, together with calibrated camera parameters and a robot model,
to reconstruct the tools, target objects, and their spatial arrangement in
simulation. Recorded motions anchor the reconstruction
to the demonstrated interaction. The resulting environment enables the agent to
replace a tool, change its functional parameters, adjust object positions, and
execute candidate action sequences while observing contact and task outcomes. The
reconstructed environment enables the agent to rapidly test candidate functional
regions through simulated tool interactions.

\subsection{Autonomous Affordance Discovery}
\label{sec:agent}

\paragraph{From examples to extraction skills.}
As shown in Fig.~\ref{fig:agentloop}(a), the agent compares source-tool demonstrations to identify the functional region
and properties that enable the task, such as position, local orientation, and
task-specific direction. It develops executable affordance-extraction skills for
these task-relevant features. Each skill extracts and encodes features as
$\mathbf f_t$ at each demonstrated time step, with spatial features expressed
in the robot-base frame through tool attachment and kinematics. Ordering these
features in time yields the demonstrated affordance trajectory.

\paragraph{Affordance validation in simulation.}
The agent validates each tool's extracted affordance features in simulation.
Specifically, it tests whether feature-based trajectory retargeting enables
different tools to complete the task. Let
$\mathbf Q=(\mathbf q_0,\ldots,\mathbf q_{T-1})$ be a feasible $T$-step robot
trajectory that completes the task with the source tool in simulation, yielding
reference features $\mathbf f_t^{\mathrm{src}}$.
For the target-tool trajectory $\mathbf Q'=(\mathbf q'_0,\ldots,\mathbf q'_{T-1})$,
features $\mathbf f^{\mathrm{tgt}}(\mathbf q'_t)$ express local functional-region
geometry in the robot-base frame, combining robot forward kinematics at
$\mathbf q'_t$ with the tool's pose relative to the wrist.
Subject to reachability and collision constraints, it minimizes
\begin{equation}
 \begin{split}
 \mathbf Q^*=\arg\min_{\mathbf Q'}\; &
 \sum_{t=0}^{T-1}\mathcal L_{\mathrm{align}}\!\left(
 \mathbf f^{\mathrm{tgt}}(\mathbf q'_t),\mathbf f_t^{\mathrm{src}}\right)
 \\ &{}+\mathcal L_{\mathrm{smooth}}(\mathbf Q').
 \end{split}
 \label{eq:retarget}
\end{equation}
Here $\mathcal L_{\mathrm{align}}$ aligns functional-region poses across source and target tools,
matching position and task-relevant orientation.
$\mathcal L_{\mathrm{smooth}}(\mathbf Q')$ is a trajectory-smoothing function.

\begin{figure}[tbp]
\centering
\includegraphics[width=\linewidth]{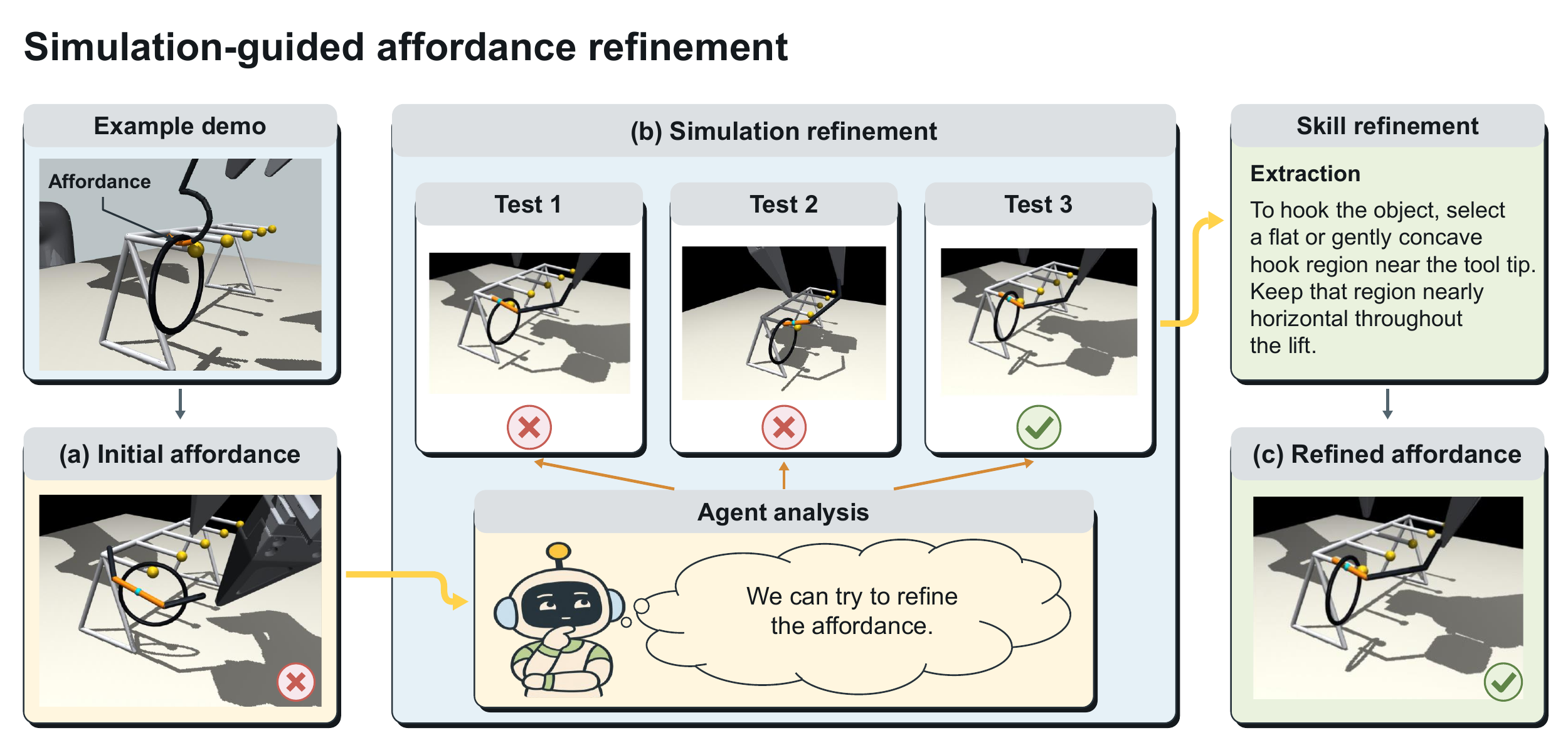}
\caption{Simulation-guided affordance refinement. In the source-tool demonstration, the
nearly horizontal hook region at the base of the hook supports the ring (orange). (a) The initial region
on the target tool misses the ring. (b) The agent tests alternative contact locations
and orientations. In Test~1, the selected region fails to enter the ring; in Test~2, the
forward-tilted hook region allows the ring to slide off during lifting; Test~3 retains the ring.
(c) The refined skill selects a flat or gently concave hook region near the tool
tip and keeps it nearly horizontal throughout the lift.}
\label{fig:sim_loop_qualitative}
\end{figure}

\paragraph{Refining and retaining skills.}
Simulation returns task outcomes, contact information, and diagnostic views.
The agent uses this feedback to refine functional regions, geometric features,
or affordance-extraction skills, then repeats the test.
Source-tool analysis determines the affordance features before VLA training;
their values follow a unified task-level encoding protocol. Successful extraction
skills and their applicability conditions enter a skill library.
The retained skills annotate all source-tool teleoperated demonstrations with tool
affordance features for VLA training.
For unseen tools, the agent retrieves and applies these skills to identify
corresponding functional regions and extract the same features. Figure~\ref{fig:sim_loop_qualitative} illustrates
how simulation tests guide revisions to candidate functional regions and orientations.

\subsection{Affordance-Conditioned VLA}
\label{sec:policy}

We condition a $\pi_{0.5}$ backbone~\cite{pi05} on the encoded affordance features
$\mathbf f_t$, current observation $\mathbf o_t=(I_t,\mathbf q_t)$, and task
instruction $\ell$. An action prediction head and an auxiliary affordance prediction
head output the action chunk and future affordance trajectory:
\begin{equation}
 \bigl(\widehat{\mathbf A}_t,\widehat{\mathbf F}_t\bigr)
 =\pi_\theta(\mathbf o_t,\ell,\mathbf f_t).
 \label{eq:policy}
\end{equation}
The training target is
$\mathbf F_t=(\mathbf f_{t+1},\ldots,\mathbf f_{t+H})$, extracted from
demonstrated action-target configurations by the agent's validated affordance-extraction skill.
Each $\mathbf f_{t+h+1}$ describes the functional region at the target of
$\mathbf a_{t+h}$, for $h=0,\ldots,H-1$. Thus $\mathbf f_t$ is the current
affordance feature vector, while $\mathbf F_t$ contains the $H$ subsequent target-step
features, paired one-to-one with the $H$ commands in $\mathbf A_t$.
Hats denote predictions.
The two heads are trained jointly with
$\mathcal L=\mathcal L_{\mathrm{act}}+\mathcal L_{\mathrm{aff}}$.
The action term is the backbone's flow-matching objective. The auxiliary term
supervises the predicted affordance trajectory, encouraging similar affordance
trajectories across tools performing the same task.
See Fig.~\ref{fig:policy}(b) and Appendix~\ref{app:model_architecture} for details.

To strengthen the VLA's ability to follow affordance guidance, we additionally
include 10\% counterfactual trajectory-retargeting data. We randomly select
chunks from source-tool demonstrations, modify the pose of the functional-region
affordance, and replan the subsequent action sequence so that the modified
affordance pose smoothly returns to the affordance-pose trajectory of the
original task demonstration. This process aims to train the VLA to execute the task
by strictly following the current affordance pose and observations of the target
object. More formally, for a sampled rigid pose perturbation
$\Delta T\in\mathrm{SE}(3)$, we construct
\begin{equation}
 \begin{aligned}
 \widetilde{\mathbf f}_t&=\Delta T\cdot\mathbf f_t,\\
 \widetilde{\mathbf A}_t
   &=\operatorname{Retarget}(\widetilde{\mathbf f}_t,\mathbf F_t).
 \end{aligned}
 \label{eq:counterfactual}
\end{equation}
Here $\Delta T\cdot\mathbf f_t$ rigidly transforms the feature's spatial
components, and $\operatorname{Retarget}$ replans from the perturbed affordance
using the original $\mathbf F_t$ as the reference, with the alignment and
smoothness objective in Eq.~\eqref{eq:retarget}. When using an unseen tool,
the policy can thus use the tool affordances analyzed
by the agent to execute the task with logic similar to the source-tool
demonstrations.

\subsection{Unseen-Tool Adaptation}
\label{sec:deployment}

At deployment, the agent first reconstructs the unseen tool in simulation, then extracts
candidate affordance parameters with its affordance-extraction skills. Using these
parameters, it retargets demonstrations for simulation tests and iteratively refines
the parameters from the feedback. The refined parameters are then encoded as
$\mathbf f_t$ using the unified task-level encoding protocol and supplied to the VLA.
Following the backbone's inference procedure, the VLA conditions on $\mathbf f_t$
and current observation $\mathbf o_t$, predicts an action chunk, executes part of
it, and receives a new observation before predicting the next chunk.
Spatial components of $\mathbf f_t$ track the robot configuration and tool attachment pose.
The agent handles tool analysis and simulation tests before real-robot execution;
the VLA controls the subsequent real-robot actions.

\section{Experiments}
\label{sec:experiments}

\begin{figure}[tbp]
\centering
\includegraphics[width=\linewidth]{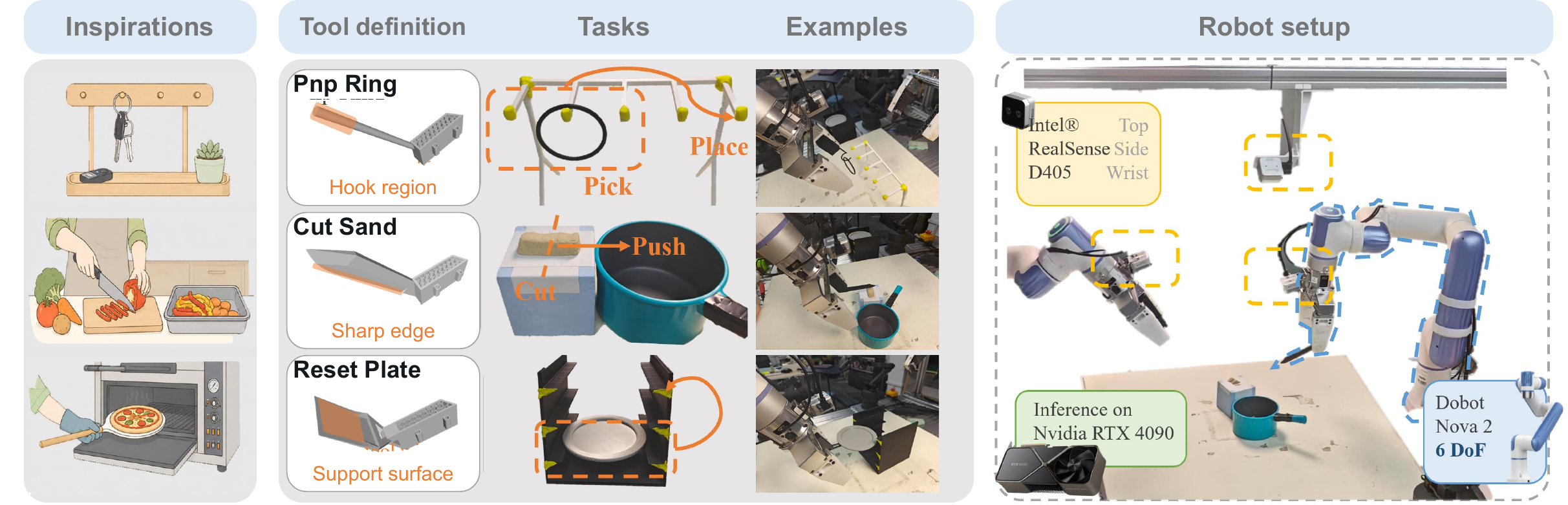}
\caption{Tool-use tasks and real-robot setup. The three tasks evaluate cross-tool transfer
in hooking, cutting, and object lifting. The leftmost column shows their everyday
inspirations, and the corresponding functional regions are highlighted in orange.
Real-robot experiments use a Dobot Nova 2 bimanual platform with RGB-D observations.}
\label{fig:apparatus}
\end{figure}

We assess the reliability and reusability of affordances discovered through autonomous
simulation experiments, as well as their effectiveness in supporting real-robot task
transfer to unseen tools. We evaluate our method on three common tool-use tasks. In \emph{Pnp Ring},
a tool moves a keyring from the left peg to the rightmost peg; \emph{Cut Sand} requires slicing
through the sand with the tool's functional edge and placing the cut portion in the
adjacent bowl; and \emph{Reset Plate} requires
retrieving a plate and placing it on the top level.

\subsection{Evaluation Setup}
\label{sec:experimental_setup}

For each task, we provide real-robot teleoperated demonstrations using a set of
\emph{source tools}. A disjoint set of \emph{unseen tools} is supplied through RGB-D inspection
observations, with no manipulation demonstrations. CAD-assisted simulation baselines additionally receive the
geometric inputs specified below. Our agent starts from the skill library built using source tools.
Real-robot evaluation tests both source tools used for demonstrations
(Table~\ref{tab:source_success}) and unseen tools (Table~\ref{tab:main_success}).
For each task and tool set, VLA-based methods use 50 evaluation episodes and Agent-Only
direct control uses 20. Each episode samples a tool uniformly from the corresponding
set. Task-specific counts and success criteria are listed in
Appendix~\ref{app:task_statistics}
(Tables~\ref{tab:task_statistics} and~\ref{tab:task_success_criteria}).

\paragraph{Simulation comparisons.}
All simulation scenes and tools are reconstructed by GPT-6 Astra from real demonstrations and
checked using trajectory replay. Based on the affordances inferred by each method, we retarget
the source-tool demonstration trajectory to unseen tools and execute the resulting trajectory
in the reconstructed simulation scene. Task success measures the reliability
of the inferred affordances. Comparisons include direct GPT-6 Astra affordance prediction
without an agentic harness (Raw VLM), MAGIC~\cite{magic}, Interaction Warping (IW)~\cite{interactionwarping}, MimicFunc~\cite{mimicfunc},
\method, R-NDF~\cite{rndf}, and TAX-Pose~\cite{taxpose}.
Where the task and output interfaces permit, methods retain their native affordance
representations and corresponding transfer procedures. Evaluation episodes use the same simulation initial
conditions and trajectory-planning program for all methods.
The IW and R-NDF rows additionally use original tool CAD, including mounting geometry,
with model matching and rigid registration assisted by the provided reconstructed proxies.
Table~\ref{tab:simulation} summarizes this comparison.

\begin{table}[t]
\centering
\caption{Simulation retargeting success (\%) with shared retargeting and robot control. We evaluate 80, 60, and 80 episodes for Cut Sand, Reset Plate, and Pnp Ring, respectively, sampling unseen tools uniformly in each episode. IW and R-NDF additionally receive original CAD with proxy-assisted registration (w/ CAD). Missing mappings and unreachable plans count as failures. Mean averages the three task rates. Raw VLM directly predicts affordances with GPT-6 Astra.}
\label{tab:simulation}
\small
\setlength{\tabcolsep}{4pt}
\renewcommand{\arraystretch}{1.02}
\begin{tabularx}{\linewidth}{@{}Lcccc@{}}
\toprule
Method & Pnp Ring & Cut Sand & Reset Plate & Mean \\
\midrule
MAGIC & 8.8 & 42.5 & 0.0 & 17.1 \\
IW w/ CAD & 3.8 & 21.3 & 0.0 & 8.3 \\
MimicFunc & 25.0 & 15.0 & 0.0 & 13.3 \\
R-NDF w/ CAD & 13.8 & 0.0 & 30.0 & 14.6 \\
TAX-Pose & 22.5 & 25.0 & 33.3 & 26.9 \\
Raw VLM & 31.3 & 43.8 & 38.3 & 37.8 \\
\textbf{Full \method} & 81.3 & 93.8 & 83.3 & 86.1 \\
\bottomrule
\end{tabularx}
\end{table}

\paragraph{Real-robot comparisons.}
\label{sec:baselines}
We construct the task environments using a Dobot Nova~2 bimanual manipulation platform and
three RealSense D405 cameras (Fig.~\ref{fig:apparatus}).
For each unseen tool, our agent is allowed to build a simulation model and run
simulation tests to determine and refine its affordance before
real-robot execution.

Tables~\ref{tab:source_success} and~\ref{tab:main_success} compare $\pi_{0.5}$ without affordance guidance, Agent-Only control,
w/o Sim. Iteration (our method without simulation-guided refinement),
Human Oracle, and Full \method. VLA-based methods share the pretrained $\pi_{0.5}$ checkpoint
and training budget. Affordance-guided variants randomly initialize the affordance encoder and
auxiliary affordance prediction head. Human Oracle uses human-provided affordances during both
training and inference to distinguish poor affordance representations from limitations of VLA
execution. Agent-Only does not use the VLA; the agent directly controls the robot's wrist
end-effector from observations. All agents use GPT-6 Astra.

\begin{table}[!t]
\captionsetup{aboveskip=0pt,belowskip=4pt}
\begingroup
\centering
\caption{Real-robot task success (\%) on source tools used for demonstrations.
VLA-based variants share the pretrained $\pi_{0.5}$ backbone and training budget.
Each task uses 50 episodes for VLA methods and 20 for GPT-6 Astra direct control
(Agent-Only), sampling tools uniformly. Human Oracle uses human-provided
affordances in training and inference. Mean averages the three task rates.}
\label{tab:source_success}
\small
\setlength{\tabcolsep}{4pt}
\renewcommand{\arraystretch}{1.06}
\begin{tabularx}{\linewidth}{@{}Lcccc@{}}
\toprule
Method & Pnp Ring & Cut Sand & Reset Plate & Mean \\
\midrule
$\pi_{0.5}$ & 34 & 12 & 20 & 22.0 \\
Agent-Only & 20 & 35 & 25 & 26.7 \\
w/o Sim. Iteration & 40 & 34 & 32 & 35.3 \\
Human Oracle & 64 & 48 & 50 & 54.0 \\
\textbf{Full \method} & 52 & 42 & 44 & 46.0 \\
\bottomrule
\end{tabularx}
\par
\endgroup

\vspace{8pt}
\begingroup
\centering
\caption{Real-robot task success (\%) on unseen tools,
using the evaluation protocol of Table~\ref{tab:source_success}.
Mean averages the three task rates.}
\label{tab:main_success}
\small
\setlength{\tabcolsep}{4pt}
\renewcommand{\arraystretch}{1.06}
\begin{tabularx}{\linewidth}{@{}Lcccc@{}}
\toprule
Method & Pnp Ring & Cut Sand & Reset Plate & Mean \\
\midrule
$\pi_{0.5}$ & 10 & 0 & 4 & 4.7 \\
Agent-Only & 25 & 30 & 20 & 25.0 \\
w/o Sim. Iteration & 28 & 20 & 18 & 22.0 \\
Human Oracle & 54 & 40 & 48 & 47.3 \\
\textbf{Full \method} & 46 & 26 & 34 & 35.3 \\
\bottomrule
\end{tabularx}
\par
\endgroup

\end{table}

\subsection{Results and Analysis}
\label{sec:results}

\paragraph{Affordance reliability.}
Table~\ref{tab:simulation} evaluates affordance reliability through simulated cross-tool
retargeting. Full \method achieves 86.1\% mean success, exceeding 37.8\% for Raw VLM,
which uses GPT-6 Astra for direct affordance prediction without an agentic harness.
This improvement demonstrates the value of simulation-guided testing and refinement
of affordance-extraction skills. Compared with other tool-adaptation methods, our agent
more effectively identifies functionally similar regions for precise trajectory retargeting
when switching tools.

\paragraph{Real-robot transfer experiments.}
Tables~\ref{tab:source_success} and~\ref{tab:main_success} report real-robot success on source
and unseen tools, respectively.
Full \method achieves 46.0\% mean success on source tools, compared with 22.0\% for
$\pi_{0.5}$. When transferred from source tools to unseen tools, Full \method retains
35.3\% mean success, outperforming both $\pi_{0.5}$ and direct Agent-Only
control. Comparing w/o Sim. Iteration with Full \method further shows that
simulation-guided refinement improves affordance-guided VLA execution on the real robot.

\paragraph{ABLATIVE \& QUALITATIVE STUDIES}
\begin{figure}[tbp]
\centering
\begin{minipage}[t]{0.48\linewidth}
\vspace{0pt}
\centering
\begin{minipage}[c][62mm][c]{\linewidth}
\centering
\includegraphics[width=\linewidth,height=60mm,keepaspectratio]{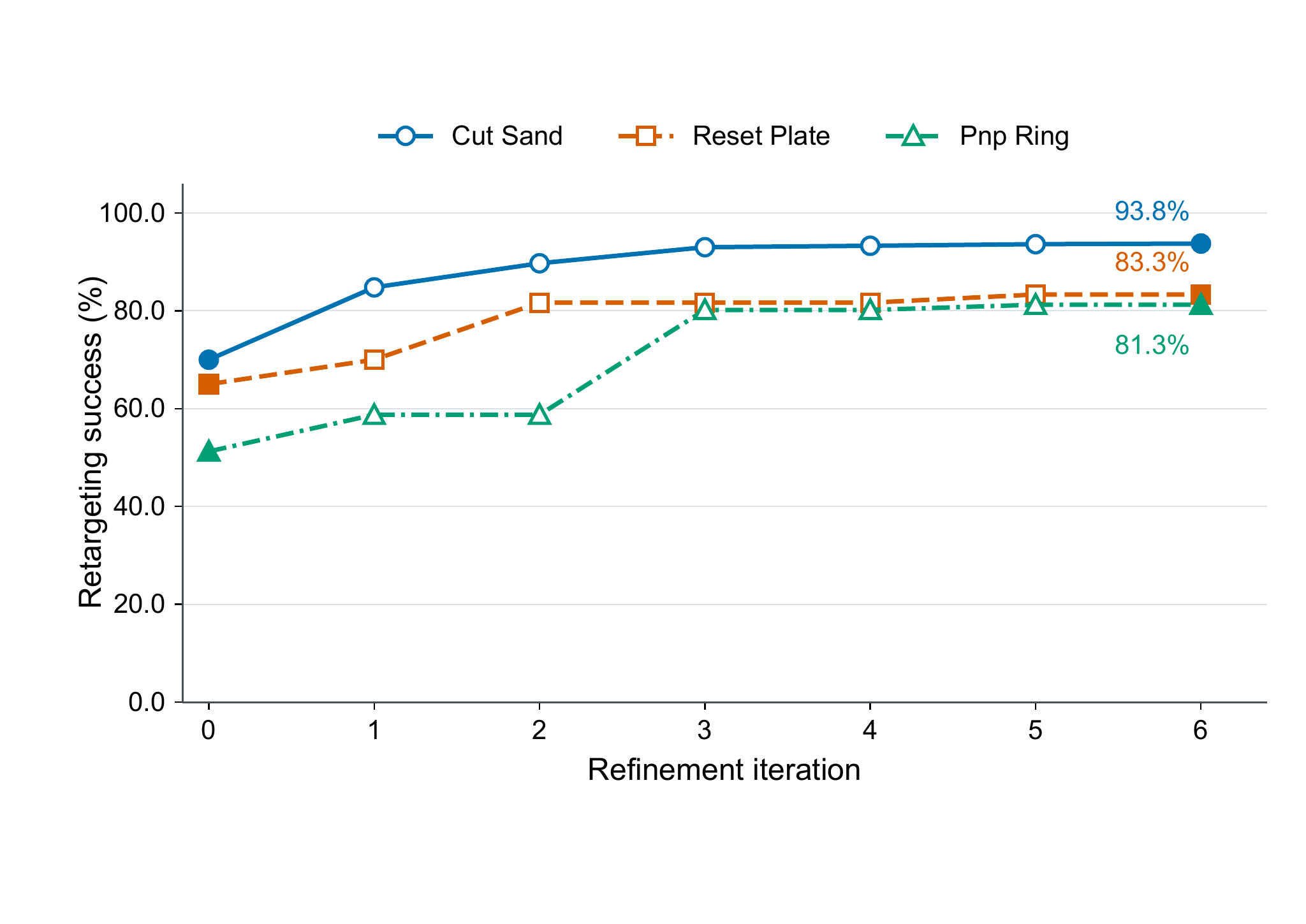}
\end{minipage}
\par\smallskip
{\small (a) Refinement iterations}
\end{minipage}\hfill
\begin{minipage}[t]{0.48\linewidth}
\vspace{0pt}
\centering
\begin{minipage}[c][62mm][c]{\linewidth}
\centering
\includegraphics[width=\linewidth,height=60mm,keepaspectratio]{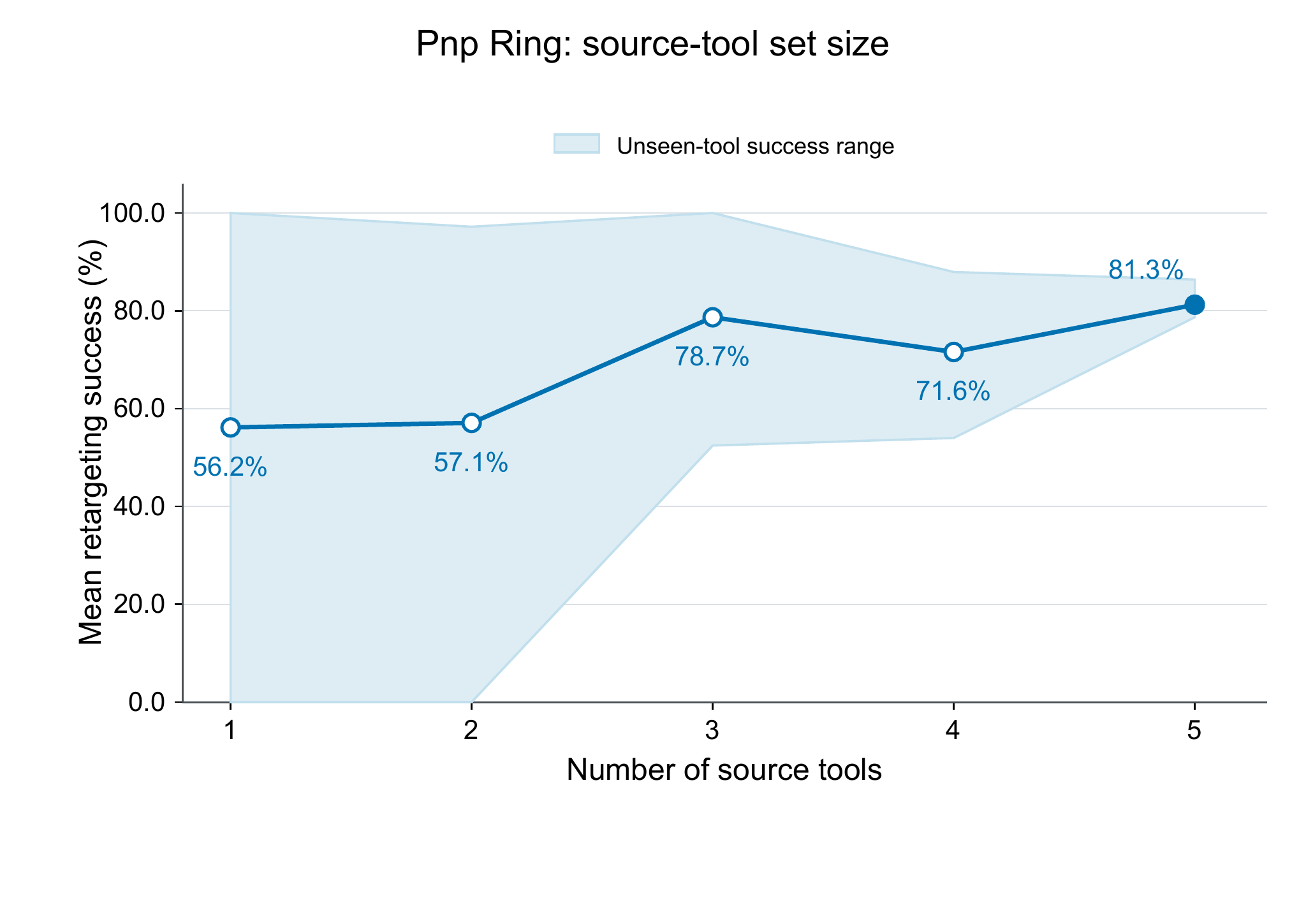}
\end{minipage}
\par\smallskip
{\small (b) Number of source tools (Pnp Ring)}
\end{minipage}
\caption{Simulation retargeting success with unseen tools.
(a) Simulation-guided refinement progressively improves task success for Cut Sand,
Reset Plate, and Pnp Ring by testing and refining tool affordances.
(b) Increasing the number of source tools for affordance discovery generally raises the mean
success rate for Pnp Ring and reduces success-rate variation across unseen tools.
The shaded band spans their minimum and maximum success rates.}
\label{fig:simulation_curves}
\end{figure}

To assess the effects of simulation-guided refinement and source-tool diversity on our
method's performance, we vary the refinement iterations and the number of source tools
used for affordance discovery.
Figure~\ref{fig:simulation_curves}(a) shows that six refinement iterations improve
retargeting success by 23.8, 18.3, and 30.0 percentage points for Cut Sand,
Reset Plate, and Pnp Ring, respectively, demonstrating the benefit of revising
affordances using simulated task outcomes.

Figure~\ref{fig:simulation_curves}(b) varies source-tool set size for Pnp Ring affordance
discovery. From one to five source tools, mean success across unseen tools rises from 56.2\% to 81.3\%,
and their minimum--maximum success-rate range narrows from 0.0--100.0\% to 78.7--86.4\%, showing
higher mean success and less variation across unseen tools. These trends suggest that with
few source tools, the agent may identify affordance features and extraction
skills suited only to a small subset of tools. Increasing source-tool diversity
helps the agent discover shared functional structure and develop more broadly
applicable affordance features and extraction skills.
Iteration-wise rates and source-tool-count results are listed in
Appendix~\ref{app:task_statistics} (Tables~\ref{tab:simulation_iterations}
and~\ref{tab:example_tool_count}).

\begin{figure}[tbp]
\centering
\captionsetup{skip=3pt}
\includegraphics[width=\linewidth]{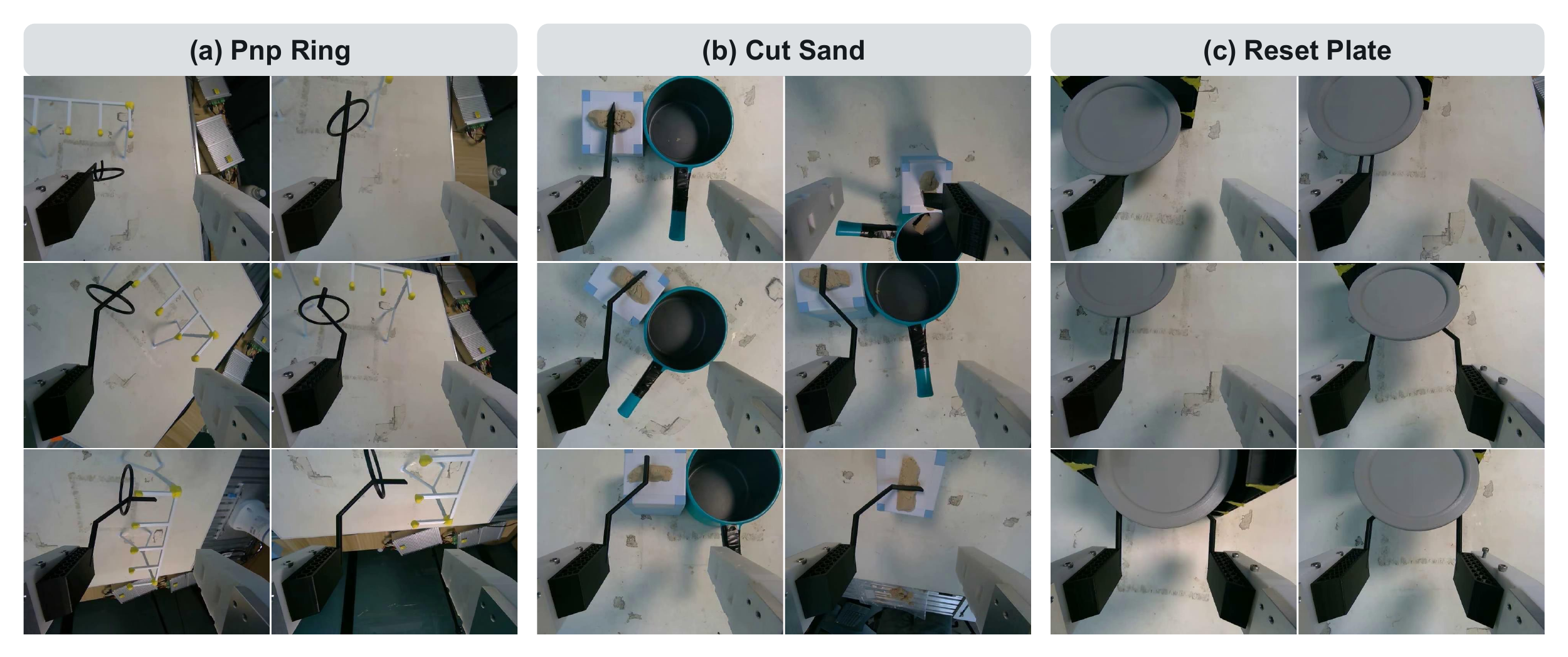}
\caption{Keyframes of real-robot task execution with different tools.
(a) \emph{Pnp Ring}: the robot uses a flat or gently concave hook region near the tool
tip to engage and lift the ring. (b) \emph{Cut Sand}: the robot uses a working edge
or terminal region to cut through the sand and transfer the cut portion
into the bowl. (c) \emph{Reset Plate}: the robot uses a suitable part of the tool
to form a support plane, lifting the plate from the rack and placing it on the upper level.
\method adapts its execution strategy to each tool's geometry.}
\label{fig:qualitative}
\end{figure}

Figure~\ref{fig:qualitative} shows keyframes of real-robot task execution with different tools.
\method adjusts its execution strategy according to each tool's geometry and functional region.

\FloatBarrier
\section{Limitations}
\label{sec:limitations}

Our method is currently limited to tool-transfer problems that can be addressed effectively
through retargeting. Although task-relevant affordance features can be determined autonomously,
some forms of tool-skill transfer in real deployment cannot be solved simply through
retargeting. In addition, the VLA may fail to follow correct affordance guidance, leading to
task failure.

\section{Conclusion}
\label{sec:conclusion}

\method uses agent reasoning and planning to guide VLA task execution by autonomously determining
task-relevant affordance features. Starting from demonstrations with a small set of source tools,
the agent reconstructs scenes, identifies functional regions and their key geometric features,
and continually improves affordance-extraction skills through simulated
retargeting tests. At deployment, the agent uses the validated skills to extract the affordances
of an unseen tool and supplies them as inputs that guide the VLA policy in completing the task.
This enables reliable skill transfer and ultimately improves task success on the real robot.

{\small
\bibliographystyle{ieeenat_fullname}
\bibliography{references}
}

\clearpage
\appendix
\numberwithin{equation}{section}
\begingroup
\setlength{\parskip}{2pt}
\section{Task-Specific Evaluation Details}
\label{app:task_statistics}
\setcounter{table}{0}
\renewcommand{\thetable}{A\arabic{table}}
\noindent\begin{minipage}{\linewidth}
\begingroup
\centering
\captionsetup{hypcap=false}
\captionof{table}{Tool and demonstration counts and real-robot evaluation protocol.
Source-tool and unseen-tool sets are evaluated separately. For each task and tool set,
VLA-based methods use 50 episodes and Agent-Only (GPT-6 Astra direct control) uses 20.
Each episode samples a tool uniformly from the corresponding set.}
\label{tab:task_statistics}
\small
\setlength{\tabcolsep}{6pt}
\renewcommand{\arraystretch}{1.12}
\begin{tabularx}{\linewidth}{@{}lCCCCC@{}}
\toprule
Task & \shortstack{Source\\tools} & \shortstack{Unseen\\tools} &
\shortstack{Demonstrations\\per source tool} &
\shortstack{Evaluation\\episodes per task\\VLA-based} &
\shortstack{Evaluation\\episodes per task\\Agent-Only} \\
\midrule
Pnp Ring & 10 & 10 & 10 & 50 & 20 \\
Cut Sand & 12 & 10 & 10 & 50 & 20 \\
Reset Plate & 5 & 5 & 10 & 50 & 20 \\
\bottomrule
\end{tabularx}
\par
\endgroup

\end{minipage}
\par\vspace{3pt}
\noindent\begin{minipage}{\linewidth}
\begingroup
\centering
\captionsetup{hypcap=false}
\captionof{table}{Task-specific success criteria used in evaluation.}
\label{tab:task_success_criteria}
\small
\setlength{\tabcolsep}{6pt}
\renewcommand{\arraystretch}{1.12}
\begin{tabularx}{\linewidth}{@{}lL@{}}
\toprule
Task & Success criterion \\
\midrule
Pnp Ring & Move the keyring to the rightmost target position. \\
Cut Sand & Slice through the sand with the tool's functional edge and place the cut portion in the adjacent bowl. \\
Reset Plate & Retrieve the plate from the rack and place it on the top-layer target. \\
\bottomrule
\end{tabularx}
\par
\endgroup

\end{minipage}
\par\vspace{3pt}
\noindent\begin{minipage}{\linewidth}
\begingroup
\centering
\captionsetup{hypcap=false}
\captionof{table}{Simulation retargeting success (\%) at each refinement iteration.}
\label{tab:simulation_iterations}
\small
\setlength{\tabcolsep}{6pt}
\renewcommand{\arraystretch}{1.08}
\begin{tabularx}{\linewidth}{@{}rCCC@{}}
\toprule
Iteration & Cut Sand & Reset Plate & Pnp Ring \\
\midrule
0 & 70.0 & 65.0 & 51.3 \\
1 & 84.8 & 70.0 & 58.8 \\
2 & 89.7 & 81.7 & 58.8 \\
3 & 93.0 & 81.7 & 80.2 \\
4 & 93.3 & 81.7 & 80.2 \\
5 & 93.6 & 83.3 & 81.3 \\
6 & 93.8 & 83.3 & 81.3 \\
\bottomrule
\end{tabularx}
\par\smallskip
\noindent Final successful episodes: 75/80 for Cut Sand, 50/60 for Reset Plate,
and 65/80 for Pnp Ring.
\par
\endgroup

\end{minipage}
\par\vspace{3pt}
\noindent\begin{minipage}{\linewidth}
\begingroup
\centering
\captionsetup{hypcap=false}
\captionof{table}{Mean and minimum--maximum range of simulation retargeting success (\%) across unseen tools for Pnp Ring
versus the number of source tools used for affordance discovery.}
\label{tab:example_tool_count}
\small
\setlength{\tabcolsep}{6pt}
\renewcommand{\arraystretch}{1.08}
\begin{tabularx}{\linewidth}{@{}rCCC@{}}
\toprule
Source tools & Mean (\%) & Minimum (\%) & Maximum (\%) \\
\midrule
1 & 56.2 & 0.0 & 100.0 \\
2 & 57.1 & 0.0 & 97.2 \\
3 & 78.7 & 52.5 & 100.0 \\
4 & 71.6 & 54.0 & 87.9 \\
5 & 81.3 & 78.7 & 86.4 \\
\bottomrule
\end{tabularx}
\par
\endgroup

\end{minipage}
\vspace{3pt}
\paragraph{Gains from simulation-guided refinement.}
From iterations 0 to 6, Table~\ref{tab:simulation_iterations} reports gains of 23.8,
18.3, and 30.0 percentage points for Cut Sand, Reset Plate, and Pnp Ring, respectively.
Pnp Ring improves most at iteration 3, and Cut Sand at iteration 1. Continued
affordance testing and refinement stabilize success at higher levels.

\paragraph{Mean success and across-tool range.}
In Table~\ref{tab:example_tool_count}, using one to five source tools for affordance discovery
raises mean success across unseen tools by 25.1 percentage points. Their minimum--maximum
success-rate range shrinks from a width of 100.0 to 7.7 percentage points, with the minimum
rising from 0.0\% to 78.7\%, showing higher mean success and less cross-tool variation.
\endgroup

\clearpage
\section{Task-Specific Affordance Discoveries}
\label{app:affordance_discoveries}

We record the agent's final discoveries for each task in two parts: the
affordance features identified through simulation-guided refinement and their
encoding. The vectors below are task-specific instances of $\mathbf f_t$;
the time index is omitted. Positions and directions are expressed in the
robot-base frame. We use $\mathbf p$ for a functional-region core, $\mathbf t$
for a unit tangent, $\mathbf n$ for a unit normal, and $\mathbf d$ for a
task-specific direction, with semantic subscripts for each task.
For a symmetric $3\times3$ matrix $\mathbf M$, the common packing operation is
\[
 \operatorname{vec}_6(\mathbf M)
 =[M_{xx},M_{yy},M_{zz},M_{xy},M_{xz},M_{yz}]^{\top}.
\]

\subsection{Pnp Ring}
\label{app:hook_affordance}

\paragraph{Discovered features.}
The agent identifies a short, locally linear terminal segment that hooks and
supports the ring. It records the region's core position $\mathbf p_{\mathrm{hook}}$,
terminal tangent $\mathbf t_{\mathrm{hook}}$, and concavity-opening direction $\mathbf d_{\mathrm{hook}}$
that allows the ring to be supported during lifting against gravity. The
directional feature is absent when the segment has no distinguishable curvature.

\paragraph{Encoding.}
We encode the terminal axis through its tangent outer product and retain the
opening direction separately:
\begin{equation}
 \mathbf f_{\mathrm{hook}}=
 \begin{bmatrix}
 \mathbf p_{\mathrm{hook}}\\
 \operatorname{vec}_6(\mathbf t_{\mathrm{hook}}\mathbf t_{\mathrm{hook}}^{\top})\\
 \mathbf d_{\mathrm{hook}}
 \end{bmatrix}\in\mathbb R^{12}.
 \label{eq:hook_affordance_encoding}
\end{equation}
An absent opening direction is stored as $\mathbf d_{\mathrm{hook}}=\mathbf 0$.

\subsection{Cut Sand}
\label{app:cut_affordance}

\paragraph{Discovered features.}
The agent identifies a short, locally linear terminal working segment. Its
features are the working-edge core $\mathbf p_{\mathrm{edge}}$, unit edge tangent
$\mathbf t_{\mathrm{edge}}$, and optional cutting-face direction $\mathbf d_{\mathrm{cut}}$
specifying how the segment cuts into the material. This directional feature is
absent for rod-like tools without a distinct sharp edge or cutting-face direction.

\begin{samepage}
\paragraph{Encoding.}
We encode the working-edge axis through the same tangent outer product and
retain the cutting-face direction:
\begin{equation}
 \mathbf f_{\mathrm{cut}}=
 \begin{bmatrix}
 \mathbf p_{\mathrm{edge}}\\
 \operatorname{vec}_6(\mathbf t_{\mathrm{edge}}\mathbf t_{\mathrm{edge}}^{\top})\\
 \mathbf d_{\mathrm{cut}}
 \end{bmatrix}\in\mathbb R^{12}.
 \label{eq:cut_affordance_encoding}
\end{equation}
\end{samepage}
\noindent An absent cutting-face direction is stored as
$\mathbf d_{\mathrm{cut}}=\mathbf 0$.

\subsection{Reset Plate}
\label{app:plate_affordance}

\paragraph{Discovered features.}
The agent selects suitable parts of the tool that together form a support plane
for lifting the plate. It records the core position $\mathbf p_{\mathrm{sup}}$ of
the selected support region, the unit normal $\mathbf n_{\mathrm{sup}}$ of the
resulting support plane, and the lifting direction $\mathbf d_{\mathrm{lift}}$.

\paragraph{Encoding.}
We encode the constructed support plane's normal axis through its outer product, using the
same six-entry packing as the line cases, and retain the signed lifting direction:
\begin{equation}
 \mathbf f_{\mathrm{plate}}=
 \begin{bmatrix}
 \mathbf p_{\mathrm{sup}}\\
 \operatorname{vec}_6(\mathbf n_{\mathrm{sup}}\mathbf n_{\mathrm{sup}}^{\top})\\
 \mathbf d_{\mathrm{lift}}
 \end{bmatrix}\in\mathbb R^{12}.
 \label{eq:plate_affordance_encoding}
\end{equation}
The six normal-axis entries are invariant to the sign of $\mathbf n_{\mathrm{sup}}$;
$\mathbf d_{\mathrm{lift}}$ retains the signed lifting direction.

\clearpage
\section{Model Architecture and Training Details}
\label{app:model_architecture}

\subsection{Single-Token Affordance Conditioning}
\label{app:affordance_token}

We retain the $\pi_{0.5}$ policy architecture~\cite{pi05}, with its Gemma-based
transformer backbone, visual encoder, language encoder, and robot-state
projection. The agent supplies the current tool's encoded functional-region feature vector
$\mathbf f_t\in\mathbb R^{12}$, constructed as described in
Appendix~\ref{app:affordance_discoveries}. A learned affordance encoder
$E_{\mathrm{aff}}$ projects this vector into a single affordance feature token
$\mathbf z_t=E_{\mathrm{aff}}(\mathbf f_t)$ at the backbone's context-embedding
width. The complete conditioning prefix is
\begin{equation}
 \mathbf X_t=
 [\mathbf X_{\mathrm{img}},\mathbf X_{\mathrm{lang}},
  \mathbf X_{\mathrm{state}},\mathbf z_t],
 \label{eq:architecture_prefix}
\end{equation}
where the first three groups encode $I_t$, $\ell$, and $\mathbf q_t$,
respectively. Thus each observation contributes one affordance feature token
alongside the standard multimodal context. Its value is determined by the
tool's current functional-region features. All twelve feature coordinates are
provided to the encoder; an absent task-specific direction is represented by the
zero vector specified in Appendix~\ref{app:affordance_discoveries}.

\subsection{Action Expert and Prediction Heads}
\label{app:prediction_heads}

The action horizon is represented by $H$ noisy action tokens in the action-expert
suffix. During each flow-matching step, the action expert processes these tokens
while attending to the conditioning prefix $\mathbf X_t$. Let
$\mathbf h_{\tau,h}$ denote its hidden state for action step $t+h$,
where $h\in\{0,\ldots,H-1\}$, at flow time $\tau$. Two heads read these states:
\begin{equation}
 \begin{aligned}
 \widehat{\mathbf v}_h
   &=D_{\mathrm{act}}(\mathbf h_{\tau,h}),\\
 \widehat{\mathbf F}_t[h]
   &=D_{\mathrm{aff}}\!\left(
     [Q_{\mathrm{aff}}(\mathbf h_{\tau,h}),\mathbf z_t]\right).
 \end{aligned}
 \label{eq:architecture_heads}
\end{equation}
The action prediction head $D_{\mathrm{act}}$ directly maps each action-step state to a
14-dimensional flow velocity. In the auxiliary affordance prediction head, a query MLP
$Q_{\mathrm{aff}}$ maps the same state to an action-step query. An output MLP
$D_{\mathrm{aff}}$ fuses this query with the single affordance feature token
$\mathbf z_t$ by concatenation and predicts a 12-dimensional affordance feature vector.
The query varies across action steps, while $\mathbf z_t$ supplies the current
tool's functional-region context throughout the chunk. Stacking these
vectors over the horizon gives $\widehat{\mathbf F}_t\in\mathbb R^{H\times12}$.
Both outputs are obtained from the same action-expert forward pass.

\subsection{Sequence Supervision and Training}
\label{app:architecture_training}

For each source tool, the agent identifies the functional region and
develops the task-relevant extraction skill. We compose the tool's pose relative
to the wrist with the wrist pose from robot
forward kinematics, expressing the functional region in the robot-base frame.
The current input $\mathbf f_t$ uses the observed configuration $\mathbf q_t$;
each subsequent target $\mathbf f_{t+h+1}$ uses the configuration commanded by
$\mathbf a_{t+h}$. Stacking these targets gives
$\mathbf F_t\in\mathbb R^{H\times12}$, paired step by step with
$\mathbf A_t\in\mathbb R^{H\times14}$.

We use the backbone's action flow-matching objective. Given Gaussian noise
$\boldsymbol\epsilon$ and flow time $\tau$, the noisy action chunk is
$\mathbf x_\tau=\tau\boldsymbol\epsilon+(1-\tau)\mathbf A_t$, with target
velocity $\boldsymbol\epsilon-\mathbf A_t$. With zero-based row indexing,
$\mathbf A_t[h]=\mathbf a_{t+h}$ and
$\mathbf F_t[h]=\mathbf f_{t+h+1}$ for $h=0,\ldots,H-1$.
The two loss terms are
\begin{equation}
 \begin{aligned}
 \mathcal L_{\mathrm{act}}
  &=\frac{1}{14H}\sum_{h=0}^{H-1}
    \left\|\widehat{\mathbf v}_h-
     (\boldsymbol\epsilon[h]-\mathbf A_t[h])\right\|_2^2,\\
 \mathcal L_{\mathrm{aff}}
  &=\frac{1}{12H}\sum_{h=0}^{H-1}
    \left\|\widehat{\mathbf F}_t[h]-\mathbf F_t[h]\right\|_2^2.
 \end{aligned}
 \label{eq:architecture_losses}
\end{equation}
The auxiliary objective supervises all twelve coordinates, including zero-valued
direction entries. The affordance encoder and auxiliary prediction head are
randomly initialized, and the VLA is initialized from the pretrained
$\pi_{0.5}$ checkpoint.
The training set includes the 10\% counterfactual trajectory-retargeting data described in
Sec.~\ref{sec:policy}.

\paragraph{Training budget.}
All VLA-based methods in the real-robot comparisons---$\pi_{0.5}$,
w/o Sim. Iteration, Human Oracle, and Full \method---are trained for 35,000 steps
with a batch size of 32.

\subsection{Policy Inference}
\label{app:architecture_inference}

At each policy update, the current observation, instruction, and affordance
feature token form the conditioning prefix. Starting from Gaussian action noise,
the action expert iteratively predicts flow velocities and denoises the action
chunk from $\tau=1$ to $\tau=0$. The robot executes part of this chunk before
receiving a new observation and replanning. The auxiliary head produces the
corresponding affordance trajectory. Spatial components of $\mathbf f_t$ are
updated from robot kinematics and the tool's attachment pose, and the encoder produces the updated feature
token for the next control cycle.

\end{document}